# Aggression-annotated Corpus of Hindi-English Code-mixed Data

## Ritesh Kumar, Aishwarya N. Reganti, Akshit Bhatia, Tushar Maheshwari

Dr. Bhim Rao Ambedkar University, Indian Institute of Information Technology-Sricity, Birla Institute of Technology and Science-Pilani, Indian Institute of Information Technology-Sricity

Agra, Sricity, Pilani, Sricity

riteshkrjnu@gmail.com, aishwarya.r14@iiits.in, akshit1811@gmail.com, tushar.m14@iiits.in

**Abstract**

As the interaction over the web has increased, incidents of aggression and related events like trolling, cyberbullying, flaming, hate speech, etc. too have increased manifold across the globe. While most of these behaviour like bullying or hate speech have predated the Internet, the reach and extent of the Internet has given these an unprecedented power and influence to affect the lives of billions of people. So it is of utmost significance and importance that some preventive measures be taken to provide safeguard to the people using the web such that the web remains a viable medium of communication and connection, in general. In this paper, we discuss the development of an aggression tagset and an annotated corpus of Hindi-English code-mixed data from two of the most popular social networking / social media platforms in India – Twitter and Facebook. The corpus is annotated using a hierarchical tagset of 3 top-level tags and 10 level 2 tags. The final dataset contains approximately 18k tweets and 21k facebook comments and is being released for further research in the field.

## 1. Introduction

In the last few years, we have witnessed a gradual shift from largely static, read-only web to quickly expanding user-generated web. There has been an exponential growth in the availability and use of online platforms where users could put their own content. A major part of these platforms include social media websites, blogs, Q&A forums and several similar platforms. All of these are almost exclusively user-generated websites. As such they change and expand at a very rapid pace. In addition to this, most of the traditionally read-only web have started to give an option to the readers to interact with the website as well as the other users by posting their comments and replying to the comments of other users.

In all of these platforms and forums, humongous amount of data is created and circulated every minute. It has been estimated that there has been an increase of approximately 25% in the number of tweets per minutes and 22% increase in the number of Facebook posts per minute in the last 3 years. It is posited that approximately 500 million tweets are sent per day, 4.3 billion Facebook messages are posted each day, more than 200 million emails are sent each day, and approximately 2 million new blog posts are created daily over the web (Schultz, 2016). We still do not have a consolidated figure on the number of comments and opinion generated on different websites but it can be safely assumed that those would be comparably staggering.

As the number of people and this interaction over the web has increased, incidents of aggression and related events like trolling, cyberbullying, flaming, hate speech, etc. too have increased manifold across the globe. While most of these like bullying or hate speech have predated the Internet, the reach and extent of Internet has given these incidents an unprecedented power and influence to affect the lives of billions of people. It has been reported that these incidents have not only created mental and psychological agony to the users of the web but has in fact forced people to deactivate their accounts and in rare instances also commit suicides. Thus the incidents of aggression and unratified verbal behaviour has not remained just a minor nuisance but has acquired the form of a major criminal activity that affects a large number of people. So it is of utmost significance and importance that some preventive measures be taken to safeguard the interests of the people using the web as well as of the web such that it remains a viable medium of communication and connection, in general.

While the initial response to handling these aggressive behaviour was to manually monitor and moderate the user-generated content, the amount and pace at which new data is being created over the web has rendered these manual methods of moderation and intervention almost completely impractical and ineffective. As such it has become imperative that such behaviours be recognised and dealt with using automatic or semi-automatic means.

However, as much as we want to deal with this automatically, it is not quite that easy to automatically recognise these, especially, using the traditional dictionary look-up or similar methods. The systems need to be more intelligent and nuanced in order to be useful in a large number of cases. Moreover, the system should also be able to recognise incidents of both overt as well as covert aggression. At the same time, it must be able to distinguish in between the ratified and unratified aggressive behavior.

## 2. Verbal Aggression

Verbal aggression could be understood as any kind of linguistic behaviour which intends to damage the social identity of the target person and lower their status and prestige (Barron and Richardson 1994, cited in Culpeper 2011). It is any kind of behaviour that upsets the social equilibrium. In general, verbal aggression can be ratified as well as unratified and it is but obvious that we are mainly concerned with the unratified aggressive behaviour. In order to build an automatic aggression-detection system, it is important that we have a good understanding of the structure of the phenomenon of aggression itself, especially the distinction between ratified and unratified behaviour such that the system flags only the most serious cases of aggression. Previous research in the field has been carried out to automatically recognise several related behaviours such as trolling (Cambria, et al., 2014; Kumar, Spezzano and Subrahmanian, 2013; Mojica, 2016; Mihaylov et al, 2015), cyberbullying (Kathick et al., 2012; Nitta et al., 2013; Dadvar, Trieschnigg and de Jong, 2014; Van Hee et al, 2015), flaming / insults (Sax, 2016; Nitin et al., 2012), abusive / offensive language (Chen et al 2012; Nobata et al., 2016) and others. However, there is hardly any theoretical insight into the structure and formation of such behaviours (some notable exceptions include Hardaker, 2010, 2013), in general, and absolutely nothing in Indian scenario. Furthermore, there is a theoretical gap in the understanding of interrelationship among these phenomena – while most of the researchers have focussed on one of these phenomena and their computational processing, it seems there is a significant overlap among these phenomenon in the way they are understood in these studies. All of these are considered undesirable, aggressive and detrimental for those on the receiving end. However, besides focussing on the intention of the initiator of these behaviours, there is hardly any discussion around their pragmatic and/or syntactic structure. So, trolling is intended "to cause disruption and/or to trigger or exacerbate conflict for the purposes of their own amusement" (Hardaker, 2010). Cyberbullying is "humiliating and slandering behavior towards other people" [Nitta et al, 2013]. Flaming intends "to offend someone through e-mail, posting, commenting or any statement using insults, swearing and hostile, intense language, trolling, etc." (Krol, 1992). Going purely by these understandings, the overlap among these phenomena are pretty obvious – as we try to classify actual data in one of these categories, the overlap becomes even more prominent. As such it might be possible to tackle all of these using similar methods, rather than building separate systems for each of these.

## 3. Data Collection

The data for the current corpus was crawled from the public Facebook Pages and Twitter. The data was mainly collected from the pages/issues that is expected to be discussed more among the Indians (and in Hindi).

For Facebook, more than 40 pages were recognised and crawled to collect the data. It included the pages of following kinds:

- News websites / organizations like NDTV, ABP News, Zee News, etc.
- Web-based forums / portals like Firstost, The Logical Indian, etc.
- Political Parties / groups like INC, BJP, etc.
- Students' organisations / groups like SFI, JNUSU, AISA, etc.
- Support and opposition groups built around recent incidents in Indian Universities of higher education like Rohith Vemula's suicide in HCU, February 9 incident in JNU, etc.

For Twitter, the data was collected using some of the popular hashtags around such contentious themes as beef ban, India vs. Pakistan cricket match, election results, opinions on movies, etc.

During collection, the data was not sampled on the basis of language and so it included data from English, Hindi as well as some other Indian languages. At a later stage, the data from languages other than Hindi and Hindi-English code-mixed were handled.

## 4. Aggression Typology

Verbal aggression could be divided into two basic types based on how it is expressed -

- Overt Aggression
- Covert Aggression

Furthermore, it can be divided into 4 different types based on the target of aggression -

- Physical Threat
- Sexual Threat / Aggression
- Identity Threat / Aggression
  - Gendered Aggression
  - Geographical Aggression
  - Political Aggression
  - Casteist Aggression
  - Communal Aggression
  - Racial Aggression
- Non-threatening Aggression

In addition to these subtypes of aggression, there two distinctions are also drawn in between aggression and abuse and aggression analysis and sentiment analysis. All of the sub categories and these two distinctions are discussed in detail in the following subsections. We also discuss the annotation scheme developed on the basis of this typology of aggression.

### 4.1 Overt Aggression

Any speech / text (henceforth, text will mean both speech as well as text) in which aggression is overtly expressed – either through the use of specific kind of lexical items or lexical features which is considered aggressive and / or certain syntactic structures is overt aggression. An example is given below -

अबे कन्हइया सुना इंसहल्लाह (लाल चड्ढी) गैंग देल्ही mcd में 31 सीट पे चुनाव लड़ा और इन्हें कुल 51 वोटों से आज़ादी मिली है। कमाल हो गया बे।

Oye Kanhaiya, I have heard that Insaallah (red chaddi) gang contested election on 31 seats in Delhi MCD and they got azadi (freedom) by a total of 51 votes. It is amazing.

## 4.2 Covert Aggression

Any text in which aggression is not overtly expressed is covert aggression. It is an indirect attack against the victim and is often packaged as (insincere) polite expressions (through the use of conventionalised polite structures), In general, lot of cases of satire, rhetorical questions, etc. may be classified as covert aggression. An example is given below -

Harish Om kya anti-national ko bail mil sakti hai? ? ?

*Harish Om can an anti-national get bail?*

## 4.3 Physical Threat

Any aggressive text that threatens to hurt the victim (an individual or a community) physically or even kill her/him can be classified as physical threat. It also includes suicide intentions, mass killings, etc. as well as potentially physically aggressive (verbal aggression transforming into physical aggression). It is potentially physically aggressive in the sense that verbal aggression might transform into actual physical aggression and as such it is essential that physical threats be recognised accurately. e.g. -

Muh kala hai dogle ka dil bhi kala gaddar hai mujhe tum dikh jaye sala juta marunga dogala deshdrohi

*This hypocrite has lost his face, his heart is also bad, as soon as I shall see you moron, I will hit you with a shoe, you hypocritical anti-national*

## 4.4 Sexual Threat / Aggression

Verbal aggression that includes graphic depiction of the actual act of sex or a threat to actually carry out these acts against the victim. e.g.

Bhosri ke jab kuchh pata nahi hai to bolta kyu hai ja ke apni gaar marwa halala me.

*You fucker, when you do not know anything then why do speak. Go and get yourself fucked in Halala.*

## 4.5 Identity Threat / Aggression

It refers to the threats to one or more of the identities of the victim. It includes aggression directed at social groups, communities, etc that the victim belongs to. It can again be of 6 kinds depending on which aspect of the identity of the victim is being attacked.

### 4.5.1 Gendered Aggression

Any text that attacks the victim because of / by referring to her/his gender. It includes homophobic and transgender attacks. It also includes attack against the victim owing to not fulfilling gender roles assigned to them or fulfilling the roles assigned to another gender. e.g.

Napushank tha Nehru... lesbo thi indira

*Nehru was impotent, Indira was a lesbian*

### 4.5.2 Geographical Aggression

Aggression aimed at the victim referring to one's place of birth / origin / living is geographical aggression. 'Geographical' in this case could imply a small area like a locality to the whole of the Earth and everything in between which one's identity is attached with. e.g.

Kahe ganda jhuth bolte ho, Sharminda h ki tum bihar se ho

*Why do you speak dirty lies. I am ashamed that you are from Bihar.*

### 4.5.3 Political Aggression

Aggression directed against the victim for her/his presumed / actual affinity / membership to a particular political group / community. It also includes aggression against the political group / community itself. e.g bjp wale jyada dhindhora pitate h hindutva ka...aur hindu me hi equality nhi de pa rhe.isliye bjp ka virodh.baki states k compare me bjp ruled state me ye jyada hota h.isliye v.

*The BJP people brag about Hindutva more than others and they are not able to ensure equality among Hindus. That is why this opposition against BJP. And also because this happens more in the BJP-ruled states.*

### 4.5.4 Casteist Aggression

Aggression aimed at the caste of the victim. e.g.

Central govt k cabinet me dekho top k 10 ya 20 ministers ko.1-2 ko chor yahi 15% wale h.

*Look in the cabinet of central government. Among the top 10 – 20 ministers, besides 1 – 2, they are only these people with 15% reservation.*

### 4.5.5 Communal Aggression

Verbal aggression towards the religious affiliation and beliefs of the victim. e.g.

Ye bhi sala gandhi khan nehru khan rahul khan jaise kisi muslim ki hi olad h

*This moron is also son of some Muslim like Gandhi Khan, Nehru Khan, Rahul Khan*

### 4.5.6 Racial Aggression

Verbal aggression aimed towards the skin color as well as ethnic identity / origin of the victim. e.g.

Aur vishal tu to indian hai.lekin tu kab se hakka noodle chinki ban gaya.indian hone ka proud hai mujhe.tu kyon Japanese ban raha hai.saale chicken noodles.

*And Vishal you are Indian. Since when have you become Hakka noodle chinki (Chinese). I am proud of being Indian. Why are you becoming Japanse. Moron chicken noodles.*

### 4.6 Non-threatening Aggression

Aggression against individual traits and choices like color of the house, choice of food (non-communal), etc. are non-threatening (even though it might still be highly distressing for the victims). It also includes most instances of personal insults, cyberbullying, etc. e.g.

तुम सिर्फ पटर पटर बोलना जानते हो और कुछ नहीं।कुछ अच्छे कर्म भी कर लिया करो गरीब दिल से दुआ देंगे सुकून मिलेगा तुझे।

*You only know how to blabber and nothing else. You do some good work also, people will bless you from their hear, you will get peace.*

### 4.7 Aggression vs. Abuse

Abuses and aggression are often correlated but neither entails the other. In cases of certain pragmatic practices like 'banter' and 'jocular mockery', abusive constructions are used for establishing inter-personal relationships and increasing solidarity. So these instances cannot be labelled as aggressive. Moreover, most of the examples that I have given above are aggressive but do not contain abuse.

However, both do co-occur in a lot of cases and lot of times we are probably more concerned with (actual) abuses (and not the banter / teasing) than aggression itself. As such, we may consider abuse/curse as one aspect of aggression (even though not strictly a sub-type of aggression). However a more in-depth analysis is needed to discover the relationship between the two.

### 4.8 Aggression Analysis vs. Sentiment Analysis

At the theoretical level, sentiment analysis seeks to analyse the psychological state of the humans through the language usage while aggression analysis only seeks to analyse the language usage without getting into the question of intentionality. On a more practical level, aggression analysis may be informed by sentiment analysis to certain extent (such that a negative sentiment may strengthen the prediction of being aggressive), it cuts across the sentiment level such that even positive sentiment could be expressed aggressively and negative sentiment need not be aggressive at all. And thus the techniques for the two may overlap at certain places but largely they will depict different sets of features and characteristics.

### 4.9 Annotation Scheme

Based on the typology discussed above, we have come up with an annotation scheme for annotating the corpus with information related to its aggression level as well as the kind(s) of aggression it exhibits. The tagset contains 3 tags at the top-level and each of the the two aggressive levels contains 2 attributes – discursive role and discursive effects (See Table 1, 2 and 3 below). Discursive effects are based on aggression typology and are of 10 kinds (with all the sub-types of aggression and abuse merged into this level). Discursive roles define the 3 roles that a person might play in an aggressive discourse and they are discussed below.

| Aggression Level | TAG | Discursive Features | |
|---|---|---|---|
| | | Discursive Role | Discursive Effect |
| Overtly Aggressive | OAG | Yes | Yes |
| Covertly Aggressive | CAG | Yes | Yes |
| Non Aggressive | NAG | May be | No |

Table 1 : Aggression Levels

| Attribute | TAG |
|---|---|
| Attack | ATK |
| Defend | DFN |
| Abet | ABT |

Table 2 : Discursive Roles

| Discursive Effect | TAG |
|---|---|
| Physical Threat | PTH |
| Sexual Aggression | SAG |
| Gendered Aggression | GAG |
| Racial Aggression | RAG |
| Communal Aggression | CoAG |
| Casteist Aggression | CaAG |
| Political Aggression | PAG |
| Geographical Aggression | GeAG |
| General Non-threatening Aggression | NtAG |
| Curse / abuse | CuAG |

Table 3 : Discursive Effects

#### 4.9.1 Discursive Roles

The three kinds of discursive roles define the role of the current post / comment in the ongoing discourse. These are defined as below -

a. **Attack**: Any comment / post which attacks a previous comment / post. It can only be aggressive. An example is given below.

मोदी जी कपड़े भी उतारवा लें तो भी हमें लगेगा कि हिन्दू रक्षा और देश हित में उतारा होगा।

*Even if Modiji would rip us in every possible way then also we would feel that it was in the interest of the Nation.*

b. **Defend**: Any comment / post which defends or counter-attacks a previous comment / post. The previous comment / post must be an attack and the current one should be in support of the victim. It could be both aggressive as well as non-aggressive. An example of each case is given below

- Kitna dukhi hai bhai tu, lagta hai teri pool kholdi kanhaiya ne. Agar tu jo ilzam uspe laga raha hai wo sach hai toh wo kiyon jail me nahi hai. **(Counter-attack)**

*How sad you are bro. It seems that Kanhaiya has shown your true face. If the accusation that you are labelling on him is true then why is he not in the prison.*

- Av tak chargesheet file nhi kar payi h Delhi police...Aur lab ne is bat Ko confirm kiya ki anti-national slogan me Kanhaiya ki aawaz nhi h.ye dusre logo no kiya.kon kiya hoga ...Samjhte hi hoge. **(Defend)**

*Delhi Police has not been able to file chargesheet till now. And the lab has confirmed that the anti-national slogan does not contain the voice of Kanhaiya. It has been done by other people. You must have an idea who has done it.*

c. **Abet**: Any comment / post which lends support and/or encourages a previous aggressive comment / post. The previous comment / post must be an attack and the current one should be in support of the aggressor. It could be both aggressive as well as non-aggressive.

Great sachchai likha aapne

*You have written great truth*

### 4.10 Annotation Conventions

Annotation is carried out at the document level – it could be a complete post, a comment or any one unit of the discourse. While annotating, the annotators were given the following instructions (in addition to a detailed annotation guidelines describing the different tags, with examples) -

- Annotators were allowed to mark more than one discursive effects if it seem that certain tweets / comments may be annotated for more than one discursive effect. Thus the annotators were asked to choose ALL the discursive effects that a comment depicts.

- A lot of tweets / comments in the data contained one or the other form of 'abuse' – in those cases the annotators were required to mark the comment as 'abuse' and also at least one more effect must be marked in such cases. So any comment will have a minimum of two effects, if it contains abuse.

- If a tweet / comment is marked as exhibiting General Non-threatening Aggression by the annotators then it cannot be marked for any other effect. In other words, any comment can be marked as General Non-threatening aggression only if it does NOT contain any other kind of threat/aggression. However, General Non-threatening aggression can also contain abuse and if it does, it should be marked so. Thus NtAG comments can be only abuse in addition to itself and nothing else.

- If the tweet / comment was in a language other than English or Hindi (or something that the annotator did not understand), it was to be marked as non-aggressive.

### 5. Inter-annotator agreement

In order to test the validity and efficiency of the above tagset, we conducted an inter-annotator agreement experiment with 4 annotators using approximately 500 test instances. Kripendorff's Alpha for this experiment was 0.49 (for the top-level annotation). Since the agreement was below par, even going by the standard of pragmatic phenomenon like aggression, we made certain changes to the annotation guidelines and conducted a second round of agreement experiments. Initially, the annotators were allowed to annotate only one discursive effect. It turned out that it resulted in a lot of disagreement among the annotators since a lot of comments / tweets could be interpreted as depicting more than one effect. Moreover, most of the discursive effects were left undefined considering that they were self-explanatory - this left a room for different interpretations by different annotators, resulting in lower inter-annotator agreement. After the first round of experiments, based on the feedback from the annotators, two major changes were done in the guidelines. The annotators were given an option to annotate the comments / tweets with multiple discursive effects. Additionally, all the categories were defined more rigorously, thereby, reducing the scope of different interpretations by the annotators. However, given the fact that aggression is a pragmatic phenomenon, the guidelines still gave annotators the flexibility of giving judgments based on their interpretation, instead of fixing the structures, lexical items, etc for each effect.

The second round of agreement experiments was done over a crowdsourcing platform, Crowdflower, using approximately 1100 test instances. In these experiments, each instance was annotated by 3 annotators. A total of 77 annotators attempted the test. It is to be noted here that each annotator did not annotate equal number of instances. The number of annotations by each annotator ranged from 135 judgments – 10 judgments. At the end of these experiments, the inter-annotator agreement for the top-level was slightly above 72%. While the agreement for the 10-class annotation of discursive effect was approximately 57%. These agreement scores were significantly higher than the scores obtained from the previous scores and so we decided to continue with the data annotation task.

### 6. Final Dataset

The complete dataset contains approximately 18k tweets and 21k facebook comments annotated with aggression level and discursive effects. The annotation was again done using the Crowdflower platform but it was done by what is known is 'internal' annotators in the Crowdflower lingo. The whole of annotation was done by 4 annotators – all of them were native speakers of Hindi, with a native-like competence in English and were pursuing a doctoral degree in Linguistics.

A preliminary study of the final annotated dataset reveals a fundamental difference in between how people communicate over Facebook and Twitter. Length-wise, approximately ⅔ of the Facebook comments are of less than 150 characters (which is approximately equal to Twitter's restriction of 140 characters). However, a comparison of the aggression level of Twitter and

Facebook (given in Figure 1 and 2) clearly shows that both the platform crucially defines the predominant aggression level – people are more vocal and overtly aggressive on Facebook in comparison to Twitter where people are more subtle and covert in expressing aggression. A different observation, however, could be made about the discursive effect where it seems a majority of the tweets as well facebook comments in the current dataset revolve around the political aggression (Figure 3). Another interesting observation could be made about the interaction between the phenomenon of code-mixing and aggression – the data shows that a majority of code-mixed comments and tweets are aggressive, while for posts in Hindi, it is equally distributed and for posts in English, it is largely non-aggressive (Figure 4).

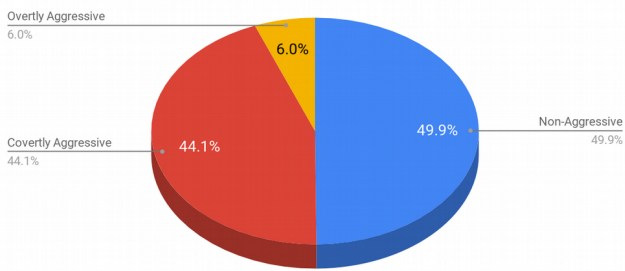

Figure 1 : Proportion of Aggressive and Non-aggressive Tweets in the dataset

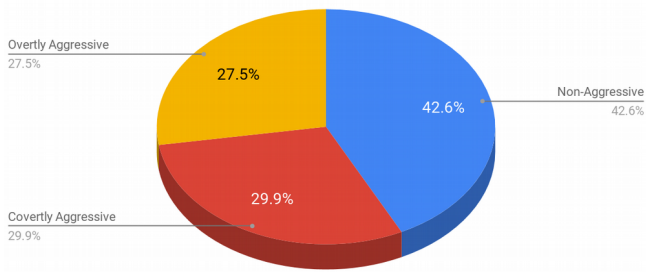

Figue 2 : Proportion of Aggressive and Non-aggressive Facebook comments in the dataset

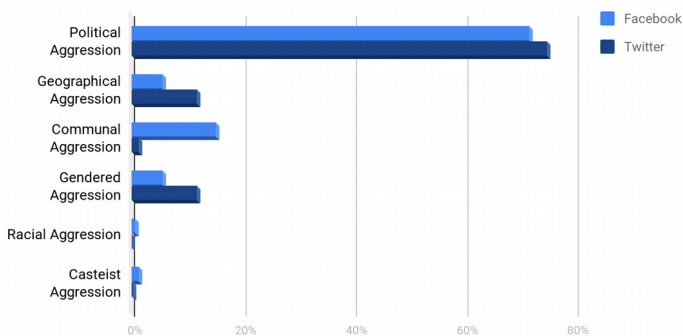

Figure 3 : Comparison of different (Identity) Discursive Effects in the dataset

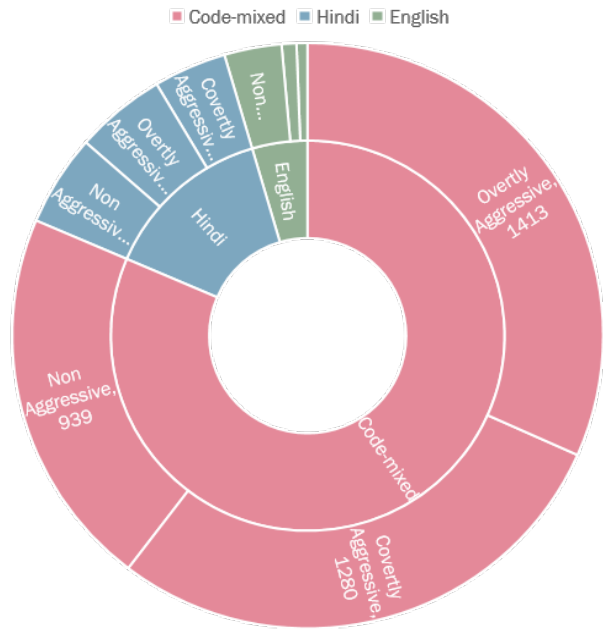

Figure 4 : Interaction of code-mixing and aggression

## 7. Summing Up

In this paper, we have discussed the development of an aggression-annotated dataset of approximately 18k tweets and 21k Facebook comments in English-Hindi code-mixed language. We have discussed the annotation scheme that was used to annotate the dataset with different levels and types of aggression. As far as we know, it is the first dataset to be annotated with different levels and kinds of aggression. We believe this dataset could be prove to be an invaluable resource for understanding as well as automatically identifying aggression and other related phenomenon like trolling and cyberbullying over the web, especially social media platforms. As such the dataset we will publicly released for free use in further research in the area.

We have recently started experimenting with the automatic identification of aggression using this dataset. However the initial results are not very encouraging with F1 score for the top-level classification - Overtly Aggressive, Covertly Aggressive and Non-aggressive - barely reaching 0.70. It shows the complexity of classifying aggression even at the most basic level and needs further investigation.

## Acknowledgements

The annotation of the corpus described in this paper is supported by an Unrestricted Research Grant from Microsoft Research India.

We are very thankful to the annotators - Deepak Alok, Mayank, Atul Ojha and Dhirendra Kumar - for their untiring efforts in finishing the annotation within a very short span of time!